\documentclass[sigconf]{acmart}
\usepackage{graphicx}
\usepackage{multirow}
\usepackage{svg}
\usepackage{bm}
\usepackage{amsmath}
\usepackage{amsfonts}
\usepackage{balance}
\usepackage{color}
\usepackage{colortbl} % 在导言区添加
\usepackage{xcolor}   % 在导言区添加
\AtBeginDocument{%
  }

\copyrightyear{2025}
\acmYear{2025}
\setcopyright{acmlicensed}
\acmConference[KDD '25]{Proceedings of the 31st ACM SIGKDD Conference on Knowledge Discovery and Data Mining V.2}{August 3--7, 2025}{Toronto, ON, Canada}
\acmBooktitle{Proceedings of the 31st ACM SIGKDD Conference on Knowledge Discovery and Data Mining V.2 (KDD '25), August 3--7, 2025, Toronto, ON, Canada}
\acmDOI{10.1145/3711896.3737115}
\acmISBN{979-8-4007-1454-2/2025/08}

\makeatletter
\renewcommand{\@shortauthors}{Yifei Li et al.}
\makeatother

\begin{document}

\title{SAGE: Scale-Aware Gradual Evolution for Continual Knowledge Graph Embedding}

\author{Yifei Li}
\orcid{0009-0005-6854-6514}
\affiliation{
  \institution{Xi'an Jiaotong University}
  \department{School of Computer Science and Technology}
  \city{Xi'an}
  \state{Shaanxi}
  \country{China}
}
\authornotemark[2]
\email{yifeilee@stu.xjtu.edu.cn}

\author{Lingling Zhang}
\orcid{0000-0003-3074-8565}
\affiliation{%
  \institution{Xi'an Jiaotong University}
  \department{School of Computer Science and Technology}
  \city{Xi'an}
  \state{Shaanxi}
  \country{China}
}
\authornote{Corresponding Author}
\email{zhanglling@xjtu.edu.cn}

\author{Hang Yan}
\orcid{0009-0009-5751-7469}
\affiliation{%
  \institution{Xi'an Jiaotong University}
  \department{School of Computer Science and Technology}
  \city{Xi'an}
  \state{Shaanxi}
  \country{China}
}
\email{hyan@stu.xjtu.edu.cn}

\author{Tianzhe Zhao}
\orcid{0000-0002-2879-2703}
\affiliation{%
  \institution{Xi'an Jiaotong University}
  \department{School of Computer Science and Technology}
  \city{Xi'an}
  \state{Shaanxi}
  \country{China}
}
\email{ztz8758@foxmail.com}

\author{Zihan Ma}
\orcid{0009-0002-2696-4943}
\affiliation{%
  \institution{Xi'an Jiaotong University}
  \department{School of Computer Science and Technology}
  \city{Xi'an}
  \state{Shaanxi}
  \country{China}
}
\email{mazihan880@stu.xjtu.edu.cn}

\author{Muye Huang}
\orcid{0009-0002-8128-3534}
\affiliation{%
  \institution{Xi'an Jiaotong University}
  \department{School of Computer Science and Technology}
  \city{Xi'an}
  \state{Shaanxi}
  \country{China}
}
\authornote{ Zhongguancun Academy, Beijing, 100094, China}
\email{huangmuye@stu.xjtu.edu.cn}

\author{Jun Liu}
\orcid{0000-0002-6004-0675}
\affiliation{%
  \institution{Xi'an Jiaotong University}
  \department{School of Computer Science and Technology}
  \city{Xi'an}
  \state{Shaanxi}
  \country{China}
}

\email{liukeen@xjtu.edu.cn}

% 简短版
% \thanks{
% * Corresponding Author\\
% † Zhongguancun Academy, Beijing, 100094, China
% }

% \thanks{
% $^1$ Ministry of Education Key Laboratory of Intelligent Networks and Network Security, Xi'an Jiaotong University, Xi’an, 710049, China.\\
% $^2$ Shaanxi Province Key Laboratory of Big Data Knowledge Engineering, Xi'an Jiaotong University, Xi’an, 710049, China.\\
% $^3$ Zhongguancun Academy, Beijing, 100094, China. * Corresponding author
% }

\begin{abstract}
% Traditional knowledge graph (KG) embedding methods aim to represent entities and relations in a low-dimensional space, primarily focusing on static graphs. However, real-world KGs are dynamically evolving with the constant addition of entities, relations and facts. To address such dynamic nature of KGs, several continual knowledge graph embedding (CKGE) methods have been developed to efficiently update KG embeddings to accommodate new facts while maintaining learned knowledge. As KGs grow at different rates and scales in real-world scenarios, existing CKGE methods often fail to consider the varying scales of updates and lack systematic evaluation throughout the entire update process. In this paper, we propose SAGE, a scale-aware gradual evolution framework for CKGE. 
% Specifically, SAGE firstly determine the embedding dimensions based on the update scales and expand the embedding space accordingly. 
% The Dynamic Distillation mechanism is further employed to balance the preservation of learned knowledge and the incorporation of new facts. 
% We conduct extensive experiments on seven benchmarks, and the results show that SAGE consistently outperforms existing baselines, with a notable improvement of 1.38\% in MRR, 1.25\% in H@1 and 1.6\% in H@10. 
% Furthermore, experiments comparing SAGE with methods using fixed embedding dimensions show that SAGE achieves optimal performance on every snapshot, demonstrating the importance of adaptive embedding dimensions in CKGE.

Traditional knowledge graph (KG) embedding methods aim to represent entities and relations in a low-dimensional space, primarily focusing on static graphs. However, real-world KGs are dynamically evolving with the constant addition of entities, relations and facts. To address such dynamic nature of KGs, several continual knowledge graph embedding (CKGE) methods have been developed to efficiently update KG embeddings to accommodate new facts while maintaining learned knowledge. Nonetheless, most existing methods treat updates uniformly, as KGs grow at different rates and scales in real-world applications, these methods often fail to consider the varying scales of updates and lack systematic evaluation throughout the entire update process.
In this paper, we propose SAGE, a scale-aware gradual evolution framework for CKGE. Specifically, SAGE firstly determines the embedding dimensions based on the update scales and expands the embedding space accordingly. The Dynamic Distillation mechanism is further employed to balance the preservation of learned knowledge and the incorporation of new facts, enabling stable knowledge evolution over time. We conduct extensive experiments on seven benchmarks, and the results show that SAGE consistently outperforms existing baselines, with a notable improvement of 1.38\% in MRR, 1.25\% in H@1 and 1.6\% in H@10. Furthermore, experiments comparing SAGE with methods using fixed embedding dimensions show that SAGE achieves optimal performance on every snapshot, demonstrating the importance of adaptive embedding dimensions in CKGE.

\end{abstract}


% \ccsdesc[500]{Information systems~Entity relationship models}

% \maketitle

% \vspace{-2mm}
% \noindent\textbf{Keywords:} Continual Knowledge Graph Embedding; Evolving Knowledge Graphs; Adaptive Scaling.
% \vspace{1mm}

\vspace{-1mm}


\vspace{-1mm}

\maketitle

\vspace{-2.5mm}

\section{Introduction}
\label{sec:intro}
Knowledge Graph Embedding (KGE) encodes entities and relations in knowledge graphs (KGs) as low-dimensional vectors, preserving their semantics~\cite{bordes2013translating, rossi2021knowledge}. These embeddings enable various KG-based applications, such as question answering~\cite{DBLP:conf/www/DongZHDTJ23}, recommendation~\cite{DBLP:conf/kdd/Wang00LC19}, and semantic search~\cite{DBLP:conf/www/XiongPC17}. However, most KGE methods only focus on static KGs, assuming a fixed graph structure for training, while real-world KGs are dynamically evolving as new entities, relations, and facts are added. For example, DBpedia, a wildely-used KG, has added more than 1 million entities, 2,000 relations, and 20 million triples between 2016 and 2018~\cite{lehmann2015dbpedia}. Such dynamism challenges traditional KGE methods, which require costly and impractical retraining from scratch to handle frequent updates in evolving KGs.

To overcome such limitations, continual knowledge graph embedding (CKGE) has emerged to incrementally update KG embeddings by integrating newly added knowledge while preserving learned representations~\cite{lkge,incde,fastkge}. Recent advances in CKGE predominantly focus on mitigating catastrophic forgetting, a persistent training dilemma wherein the acquisition of novel knowledge substantially undermines the integrity of learned representations, through various effective strategies like embedding transfer~\cite{lkge}, knowledge distillation~\cite{incde}, and parameter-efficient optimization~\cite{fastkge}. These methods effectively enhance computational efficiency and maintain semantic consistency across KG updates, avoiding retraining from scratch.

\begin{figure}[t]
  \centering
  \includegraphics[width=0.95\linewidth]{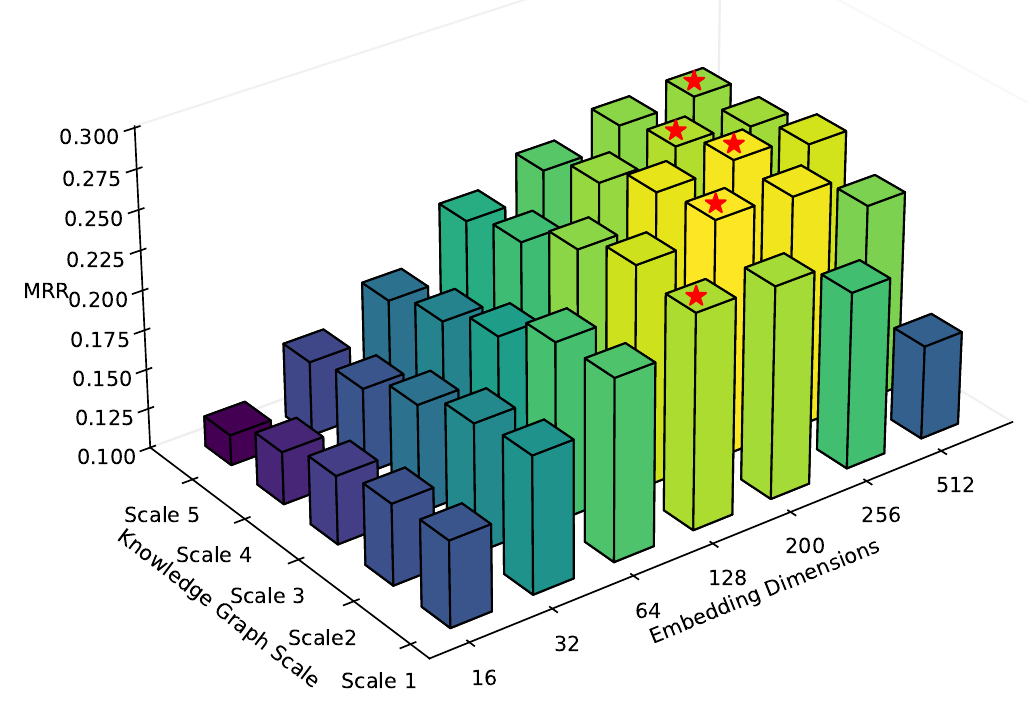}
  \caption{Effect of embedding dimensions on different KG scales. The 3D histogram visualises the embedding performance across different dimensions and scales, with red stars marking the maximum MRR value along the embedding dimensions for each scale.}
  \label{fig1}
  \vspace{-0.6cm}
\end{figure}

Despite their effectiveness, existing methods merely consider the importance of individual facts but ignore the impact of varying update scales of evolving KGs, which can range from minor changes to substantial growth. 
For instance, in an e-commerce platform, the applied KGs may either grow in frequent but small-scale updates, such as the daily addition of new users, or evolve in large-scale changes when incorporating entirely new marketplaces.
This diversity in update scales necessitates CKGE methods to possess dynamic adaptability while ensuring robust preservation of learned knowledge, thereby introducing additional challenges.
Motivated by the scaling laws in deep learning 
which demonstrate that an increase in the number of model parameters requires more data to achieve better performance~\cite{kaplan2020scaling, zhang2024scaling, droppo2021scaling}, 
we investigate the potential connection between KG scale and embedding dimensions. 
Our preliminary experiments, as illustrated in Figure~\ref{fig1}, reveal that the optimal embedding dimension (marked by red stars) shows a strong correlation with graph size, where achieving peak performance requires progressively larger dimensions as the graph expands. 
This empirical finding suggests the limited embedding capabilities of CKGE methods that employ fixed embedding dimensions.
Given this dimensional bottleneck, it's essential to design adaptive parameter adjustment strategies that dynamically align embedding dimensions with the evolving KG across different learning stages.
Considering existing parameter adjustment approaches, which tend to focus on stage-specific patterns when allocating parameters, such as selectively activating different parts of the model~\cite{chen2024dynamic, zhang2024fprompt}, their localized design inherently limits knowledge transfer between different learning stages. 
Thus, addressing these two interconnected challenges - \emph{adaptive dimension adjustment} and \emph{effective cross-stage knowledge transfer} - is pivotal to advance CKGE methods.

To this end, we propose an \textbf{S}cale-\textbf{A}ware \textbf{G}radual \textbf{E}volution framework, called \textbf{SAGE}, which is designed to dynamically update model parameters to accommodate the continual growth of evolving KGs. 
By leveraging scalable mechanisms and efficient data utilization, SAGE effectively balances knowledge retention and adaptation in evolving graphs.
To begin with, SAGE firstly evaluates the scale of the updated graph and generate footprints to capture the importance and effectiveness of entities and relations, guiding subsequent stages. 
Then, a lightweight dimension expansion mechanism are proposed for parameter expansion, which utilize existing embeddings as input features for the expanded dimensions. 
Combined with replaying a small subset of challenging samples, SAGE efficiently preserves semantic consistency between original and expanded representations. 
To enhance learning efficiency, training is performed exclusively on the newly added triples, guided by the footprints generated in the first step. These footprints dynamically adjust the distillation intensity for old entities and relations, enabling the model to strike a better balance between preserving prior knowledge and integrating new information.
Extensive experiments show that our method effectively retains prior knowledge, adapts to new facts, and handles large-scale evolving KGs with minimal data.

In summary, our contributions are as follows:
\begin{itemize}
    \item We explore the impact of dynamically growing parameters in CKGE tasks, demonstrating that aligning model capacity with the scale of graph updates enhances continual learning performance.
    \item We introduce SAGE, a framework that adaptively expands its model size to accommodate evolving KGs. Through its light-weighted adaptive update paradigm and two-stage update strategy, SAGE effectively maintains the balance between preserving existing knowledge and incorporating new information.
    \item Through extensive experiments on CKGE datasets with different growth patterns, SAGE achieves competitive results. In particular, SAGE consistently delivers optimal results at nearly every intermediate stage, achieving notable improvements in this aspect compared to existing approaches. 
\end{itemize}

To facilitate reproducibility, we release the source code and data at: https://doi.org/10.5281/zenodo.15520114.

% , extracting general knowledge during the reconstruction process.

% Regarding our proposed method, GPHT, it stands out as a generative self-supervised pretraining approach, distinguishing itself from existing methods in two key aspects.
\section{Definitions and Problem Statement}
\subsection{Evolving Knowledge Graph}

A knowledge graph $\mathcal{G} = (\mathcal{E}, \mathcal{R}, \mathcal{T})$ consists of entities $\mathcal{E}$, relations $\mathcal{R}$, and triples $\mathcal{T}$, where each triple $(h, r, t) \in \mathcal{T}$ , with $h, t \in \mathcal{E}$ and $r \in \mathcal{R}$, represents a factual statement.
An evolving knowledge graph represents the temporal evolution of a KG through a sequence of snapshots $\mathcal{G}_i = (\mathcal{E}_i, \mathcal{R}_i, \mathcal{T}_i)$ at the discrete time step $i$, where $\mathcal{E}_i$, $\mathcal{R}_i$, and $\mathcal{T}_i$ denote the respective sets at time $i$.
To characterize the dynamic growth between consecutive snapshots, we define $\Delta \mathcal{T}_i = \mathcal{T}_i - \mathcal{T}_{i-1}$, $\Delta \mathcal{E}_i = \mathcal{E}_i - \mathcal{E}_{i-1}$, and $\Delta \mathcal{R}_i = \mathcal{R}_i - \mathcal{R}_{i-1}$ as the sets of newly added triples, entities, and relations at time $i$, respectively.

\subsection{Continual Knowledge Graph Embedding}

Continual knowledge graph embedding (CKGE) focuses on learning and maintaining vector representations for entities and relations in an evolving knowledge graph. Given a sequence of KG snapshots ${\mathcal{G}_i}$, where each snapshot $\mathcal{G}_i = (\mathcal{E}_i, \mathcal{R}_i, \mathcal{T}_i)$, CKGE embeds entities and relations into a $d$-dimensional vector space. At each time step $i$, when new triples $\Delta \mathcal{T}_i$ appear alongside newly introduced entities $\Delta \mathcal{E}_i$ and relations $\Delta \mathcal{R}_i$, the model learns representations for these new elements while updating the embeddings of existing entities $\mathcal{E}_{i-1}$ and relations $\mathcal{R}_{i-1}$ to align with the current snapshot. The resulting embeddings are expected to effectively capture the semantic relationships among all entities $\mathcal{E}_i$ and relations $\mathcal{R}_i$, while preserving previously learned knowledge.

\section{Investigation on Dimension Expansion}
 
As introduced in Section \ref{sec:intro}, when KGs evolve and grow in scale, a fundamental challenge lies in determining appropriate embedding dimensions - insufficient dimensions may restrict the model's expressiveness, while excessive dimensions can lead to computational overhead and potential overfitting. 
In this section, instead of pursuing precise relationships, we aim to explore practical simulation for expanding dimension to the growing KG scales with flexibility and generality.

\label{sec: genTren}
\begin{figure}[!h]
  \centering
  \includegraphics[width=\linewidth]{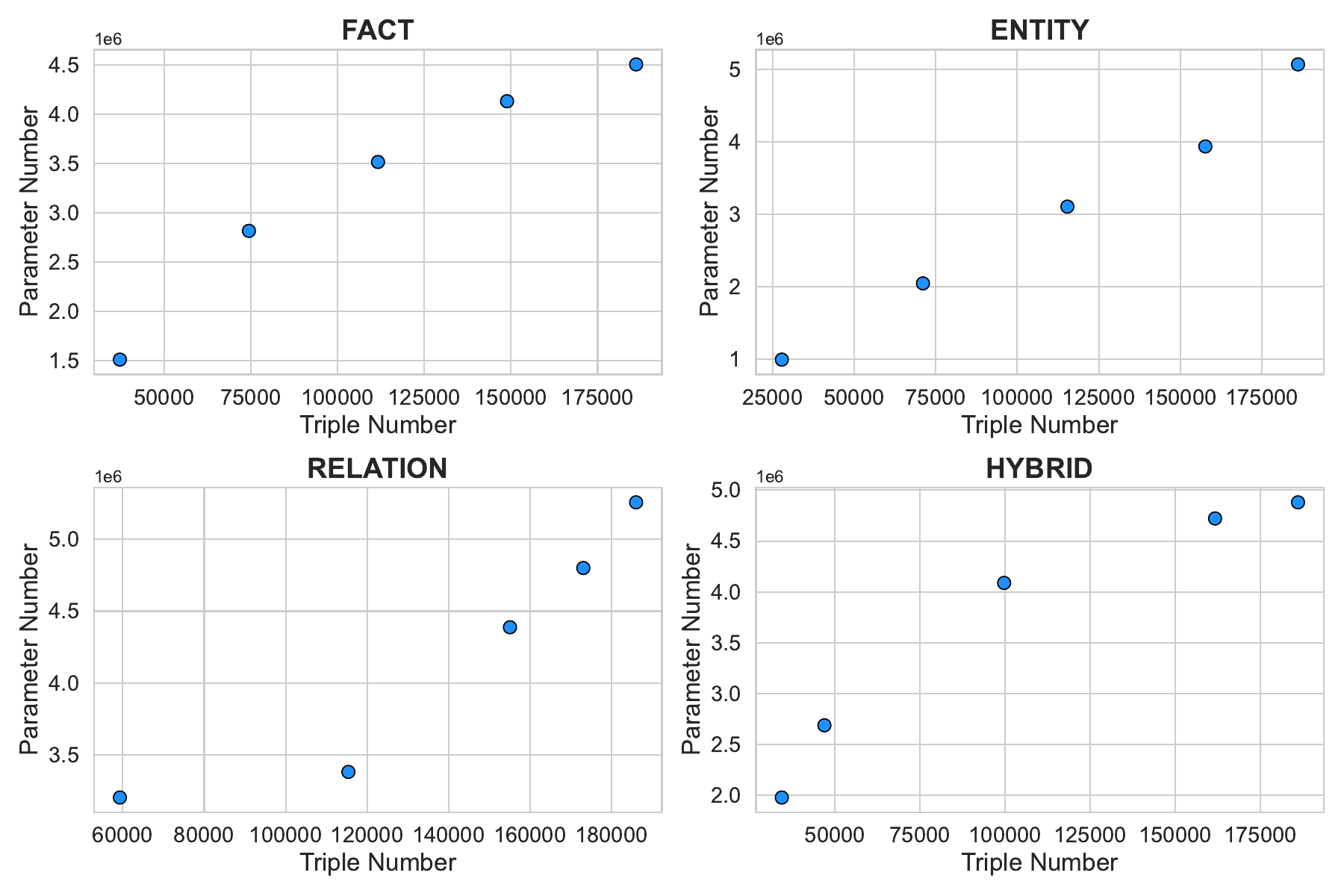}
  \vspace{-5mm} 
  \caption{Scatter plots of triple numbers (x-axis) vs. optimal parameter counts (y-axis) for four datasets: FACT, ENTITY, RELATION, and HYBRID. Each point represents the average parameter count across the optimal results obtained under different dimensions at a specific data scale.}
  \label{fig2}
\end{figure}

\vspace{-5mm}

\begin{figure}[!h]
  \centering
  \includegraphics[width=0.95\linewidth]{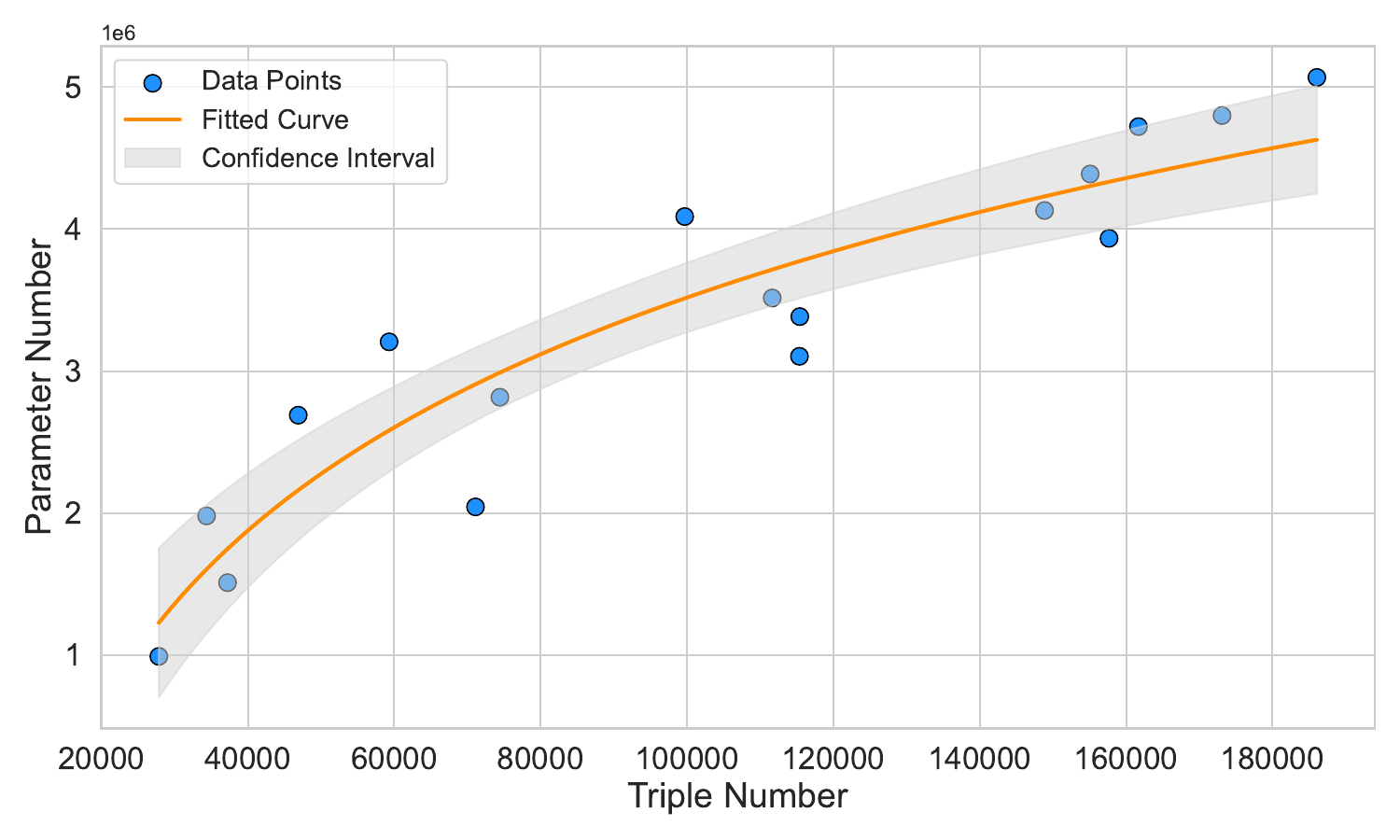}
  \vspace{-5mm}
  \caption{Logarithmic relationship between the number of triples (x-axis) and the total parameter count (y-axis) across snapshots for seven datasets with different growth strategies.}
  \label{fig3}
\end{figure}

\subsection{General Trends Across Datasets}

To explore the potential relationship between the embedding dimension and KG scales in achieving optimal performances, 
we select four representative CKGE datasets: FACT, ENTITY, RELATION, and HYBRID~\cite{lkge}, each reflecting different characteristics of KG evolution. 
For each dataset, we systematically vary the embedding size (i.e., the dimensionality of the representation space) in increments of 50 and evaluate the model's performance under different configurations. The evaluation is conducted across multiple performance metrics, and the embedding size that achieves the best overall performance is identified as the optimal configuration. At each snapshot of the dataset (corresponding to different KG scales), we determine the optimal embedding size and estimate the total parameter count by multiplying this size by the number of entities and relations in the corresponding snapshot. 

As shown in Figure~\ref{fig2}, all these datasets exhibit consistent growth patterns as they evolve over time, with the number of triples increasing from approximately 2 million to 6 million across different snapshots. Specifically, the RELATION dataset shows the steepest growth curve, while FACT, ENTITY, and HYBRID demonstrate similar growth trajectories. These patterns suggest that despite different evolution characteristics, the overall scale of knowledge graphs tends to follow predictable growth trends.

\subsection{Dimension Expansion Simulation}   

Based on the observations in Section \ref{sec: genTren}, we aggregate results across all datasets and fit them to a logarithmic curve that models the relationship between data scale ($N$) and the total number of parameters ($P$), as shown in Figure~\ref{fig3}. The fitted curve with its confidence interval demonstrates that the parameter count $P$ scales logarithmically with the graph size $N$:
\begin{equation}
P = a \cdot \log_b(N),
\end{equation}
where $a$ and $b$ are constants derived from regression. The parameter count $P$ relates to the embedding dimension $d$ through the structure of the graph: $P = d \cdot (\mathcal{|E|}+ \mathcal{|R|})$, where $\mathcal{|E|}$ and $\mathcal{|R|}$ represent the numbers of entities and relations respectively.

This logarithmic relationship reveals an interesting pattern: knowledge graphs appear to naturally accommodate efficient representations where embedding dimensions scale sublinearly with graph size. The confidence interval (shown in blue shading) suggests there exists a flexible range of suitable embedding dimensions $[d_{\text{min}}, d_{\text{max}}]$ that can effectively capture the semantic relationships at any given scale. This finding aligns with the intuition that knowledge graphs possess inherent structural regularities that enable efficient low-dimensional representations even as they grow.

In the next section, we detail how this estimation process is incorporated into a broader framework for dynamic knowledge graph representation learning.

\begin{figure*}[htbp]
  \centering
  \includegraphics[width=\linewidth]{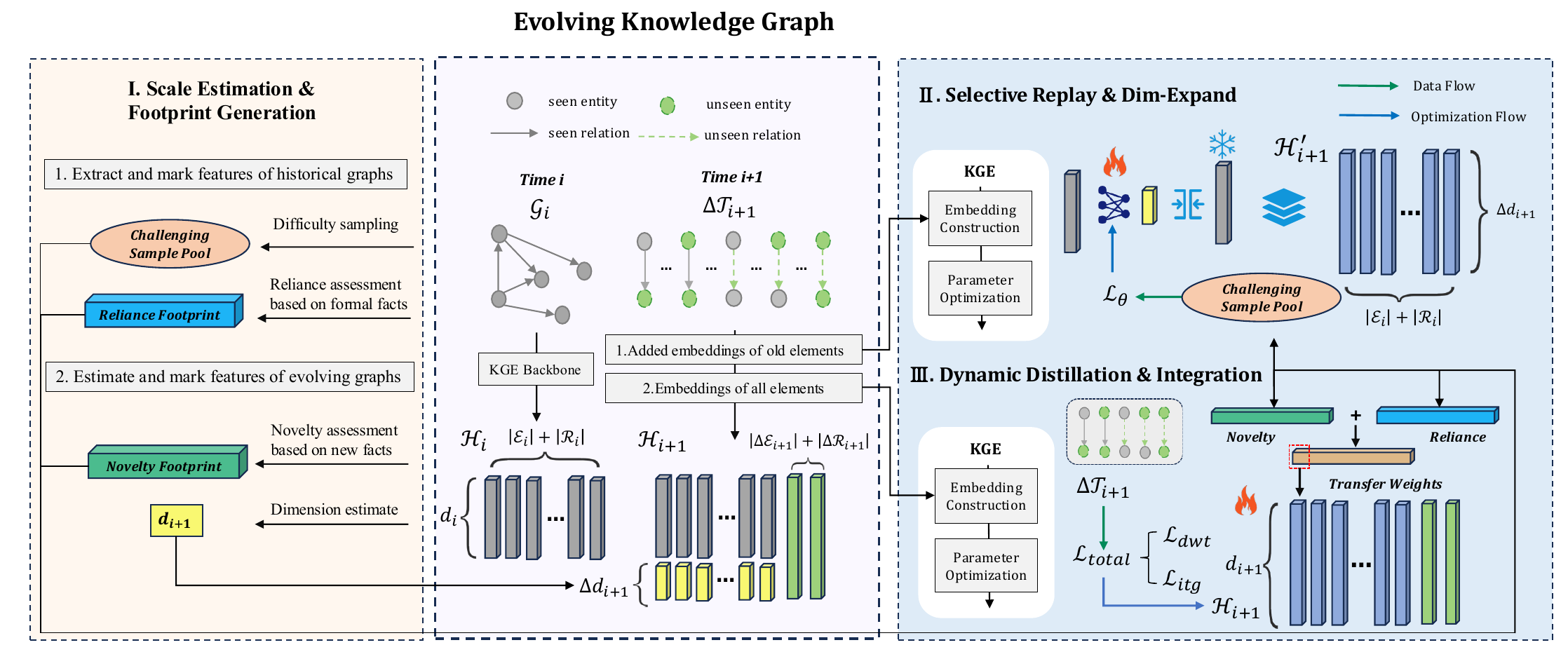}
  \caption{Overview of the SAGE framework for continual knowledge graph embedding.}
  \label{fig:model}
\end{figure*}

\section{Proposed Method}
The framework of SAGE is illustrated in Figure ~\ref{fig:model}. Overall, as a KG evolves over time, we employ dynamic parameter expansion to accommodate new knowledge while preserving existing information. 
The framework consists of three main stages: (1) Scale Estimation and Footprint Generation, (2) Selective Replay and Embedding Expansion, and (3) Dynamic Distillation and Integration.

\subsection{Scale Estimation \& Footprint Generation}\label{sec:pd}
Building upon our previous analysis of the relationship between model capacity and graph scale, we develop practical strategies for dimension estimation and knowledge preservation in evolving knowledge graphs.

\subsubsection{Adaptive Dimension Strategy}
Guided by the established logarithmic relationship between parameter requirements and graph size, we design an adaptive dimension update strategy that balances three key principles:

1) Computational tractability: We maintain stable computational overhead by employing fixed-step adjustments when dimensions are near the optimal range;

2) Sufficient initial capacity: For newly initialized or severely underfitted representations, the strategy allows rapid dimension expansion;

3) Controlled dimensional changes: Inspired by knowledge distillation in KGE ~\cite{zhu2024croppable,liu2023iterde}, we adopt a conservative approach that strictly limits the magnitude of dimension changes during updates to minimize potential information loss in knowledge transfer.

These principles are implemented through the following update rule:
\begin{equation}
d_{i+1} =
\begin{cases}
r \cdot d_i, & \text{if } r \cdot d_i \leq y_{\text{min}}, \\
y_{\text{min}}, & \text{if } \frac{y_{\text{min}}}{r} < d_i \leq y_{\text{min}}, \\
y, & \text{if } y_{\text{min}} < d_i \leq y, \\
d_i + \text{step}, & \text{if } y < d_i \leq y_{\text{max}}, \\
d_i, & \text{if } y_{\text{max}} < d_i, \\
y_{\text{max}}, & \text{otherwise}.
\end{cases}
\end{equation}
where $y_{\text{min}}$, $y$, and $y_{\text{max}}$ denote the minimum, target, and maximum bounds of the predicted dimension. The scaling factor $r$ controls the magnitude of dimensional changes, while step-wise adjustments ensure gradual convergence to optimal dimensions.

\subsubsection{Footprint Generation}
To effectively guide the learning process in evolving knowledge graphs, we need to identify the importance and learning characteristics of each knowledge element. We introduce footprints as statistical indicators that capture both the historical significance and future relevance of entities and relations.

For computational efficiency, we adopt an enhanced frequency-based approach rather than more complex measures like betweenness centrality. Our footprint design considers two key aspects: novelty in upcoming changes and historical reliance with quality assessment.

For an entity or relation $x_i \in \mathcal{G}_i$, we first compute its novelty footprint based on its occurrence in new triples:
\begin{equation}
f_n(x_i) = \left| { (h, r, t) \in \Delta \mathcal{T}_{i+1} \mid x_i \in {h, r, t}} \right|,
\end{equation}
where $f_n(x_i)$ counts how frequently $x_i$ appears in the new triple set $\Delta \mathcal{T}_{i+1}$ as either head entity, relation, or tail entity.

Furthermore, we consider both data evolution and representation quality in our footprint design. The reliance footprint is calculated as:
\begin{equation}
f_r(x_i) = \left| { (h, r, t) \in \mathcal{T}_{i} \mid x_i \in {h, r, t}} \right|\exp \left[-(1-R_i) \right],
\end{equation}
where $f_r(x_i)$ measures the occurrence of $x_i$ in existing triples $\mathcal{T}_{i}$, and $R_i$ denotes its MRR score from previous learning as a representation quality indicator.

\subsubsection{Difficulty Sampling}

To support dimension expansion, we focus on identifying challenging examples from the existing knowledge graph $\mathcal{G}_i$ where current representations show limitations. Our hypothesis is that these hard-to-model cases indicate the need for enhanced representation capacity.

We propose an entropy-based sampling mechanism to measure model uncertainty. For a sample $s$ in $\mathcal{T}_i$, we compute its entropy:
\begin{equation}
H(s) = -\sum_{j} p_j(s) \log p_j(s),
\end{equation}
where $p_j(s)$ is the predicted probability from the scoring function. The sampling probability is then formulated as:
\begin{equation}
P(s) = \frac{\exp(H(s))}{\sum_{s' \in \mathcal{T}_i} \exp(H(s'))}.
\end{equation}

Based on this distribution, we sample a challenging set $D_r$ of size $k$:
\begin{equation}
D_r = \{s'_1, s'_2, \dots, s'_k\}, \quad s'_i \sim P(s)
\end{equation}
These samples with higher uncertainty guide the subsequent dimension expansion process.

\subsection{Selective Replay and Embedding Expansion}
During dimension expansion from $d_i$ to $d_{i+1}$, we face two distinct scenarios: handling new entities $\Delta \mathcal{E}_{i+1}$ and relations $\Delta \mathcal{R}_{i+1}$, and migrating existing entities $\mathcal{E}_{i}$ and relations $\mathcal{R}_{i}$ with established representations $\mathcal{H}
_{i}$. For new elements, directly initializing parameters in the higher-dimensional space is a feasible approach. However, for historical elements with well-established representations, a systematically designed expansion strategy is crucial to preserve semantic integrity while ensuring a smooth transition to the higher-dimensional space.

We propose a knowledge-guided dimension expansion method specifically designed to facilitate the migration of existing representations. This method employs a lightweight network architecture that leverages previously learned knowledge to guide the expansion process. The network parameters are optimized using challenging samples selected during the initial phase, as detailed below:
\begin{equation}
\mathcal{H}_{i+1}^{\prime} = \bigcup_{j=1}^{|\mathcal{E}_i| + |\mathcal{R}_i|} \Big\{\text{concat}\big(\mathbf{h}_i^j, f_{\theta}(\mathbf{h}_i^j)\big)\Big\},
\end{equation}

where $\mathcal{H}_{i+1}'$ is an intermediate representation that includes updates for old entities $\mathcal{E}_i$ and relations $\mathcal{R}_i$. At this stage, no updates have been applied to new entities $\Delta \mathcal{E}_{i+1}$ and new relations $\Delta \mathcal{R}_{i+1}$.

Here, the function $f_\theta: \mathbb{R}^{d_i} \to \mathbb{R}^{d_{i+1} - d_i}$ is a learnable transformation parameterized by $\theta$, which maps the $d_i$-dimensional representation of entity or relation to $\Delta d_{i+1}$-dimensional feature space. The output of $f_\theta$ is concatenated with the original representation $\mathbf{h}_i^j \in \mathbb{R}^{d_i}$, resulting in an updated representation $\mathbf{h}_{i+1}^j \in \mathbb{R}^{d_{i+1}}$.
For fair comparison, we take TransE as the base KGE model, where the loss function for optimizing parameters $\theta$ is formulated as:
\begin{equation}
\mathcal{L}_{\theta}=\sum_{(h,r,t)\in\mathcal{D}_r}\max\left(0,\gamma+f(\mathbf{h},\mathbf{r},\mathbf{t})-f(\mathbf{h}^{\prime},\mathbf{r},\mathbf{t}^{\prime})\right)
\end{equation}
where $\gamma$ is the margin. $({h}', {r}, {t}')$ represents the embedding of a negative triplet.
This architecture offers notable advantages in both representation quality and computational efficiency. Freezing the original parameters preserves the integrity of historical knowledge, while leveraging existing knowledge as guidance ensures semantic consistency in the expanded dimensions. Compared to direct initialization methods, the reduced parameter space decreases computational overhead during expansion. Furthermore, the guided expansion approach provides a structured framework for utilizing existing knowledge representations, which can enhance stability and efficiency in learning within the expanded dimensions.

\subsection{Dynamic Distillation and Integration}
At this stage, we optimize the representations of all entities and relations involved in the newly added triples. Unlike the second stage, both new and existing representations are updated jointly.

The novelty and reliability metrics from the first stage are used to guide the updates of existing weights, while considering the number of triples from both the new and existing datasets. This ensures a balanced optimization process, where both new and existing representations are included. The integration loss function for this step is:
\begin{equation}
\mathcal{L}_{itg}=\sum_{(h,r,t)\in\Delta\mathcal{T}_i}\max\left(0,\gamma+f(\mathbf{h},\mathbf{r},\mathbf{t})-f(\mathbf{h}^{\prime},\mathbf{r},\mathbf{t}^{\prime})\right)
\end{equation}
While the formulation is similar to the earlier loss function used for optimizing $\theta$, the objectives of the two stages are fundamentally different: the former aims to expand feature dimensions by learning a transformation parameterized by $\theta$, whereas this stage directly trains the KGE model to enhance the representation quality of the newly added triples. This distinction ensures that the expanded feature space is effectively utilized, and the new knowledge is seamlessly integrated into the existing graph, thereby improving both representation quality and maintaining overall semantic consistency.

For newly added triples that involve existing entities or relations, we introduce a dynamic weight transfer mechanism to maintain the balanced inheritance of knowledge. Inspired by related works on knowledge graph embedding distillation, we formulate the loss function to dynamically regulate representation similarity as follows:
\begin{equation}
\mathcal{L}_{dwt}=\sum_{x_j\in\left(\mathcal{E}_{i}\cup\mathcal{R}_{i}\right)}\frac{f_r(x_j)}{f_n(x_j)}\|\mathbf{h}_j-\mathbf{h}_{j}^{'}\|_2^2
\end{equation}
Here, $f_n(x_j)$ and $f_r(x_j)$ represent the novelty and reliability footprints respectively. The vector $\mathbf{h}_j$ represents the updated embedding for each entity or relation, while $\mathbf{h}_j'$ denotes its intermediate representation from the second phase.
The final loss function consists of two components: the integration loss and the dynamic weight transfer loss:

\begin{equation}
\mathcal{L}=\mathcal{L}_{itg}+\alpha \mathcal{L}_{dwt}
\end{equation}

By adjusting the balance factor $\alpha$, we can flexibly control the relative importance of these two losses. This design ensures both effective learning of new knowledge and smooth transition from historical knowledge, thereby enabling continuous knowledge accumulation and updating.

\section{Experiments}

\subsection{Experimental Setup}
\begin{figure}[!t]
  \centering
  \includegraphics[width=\linewidth]{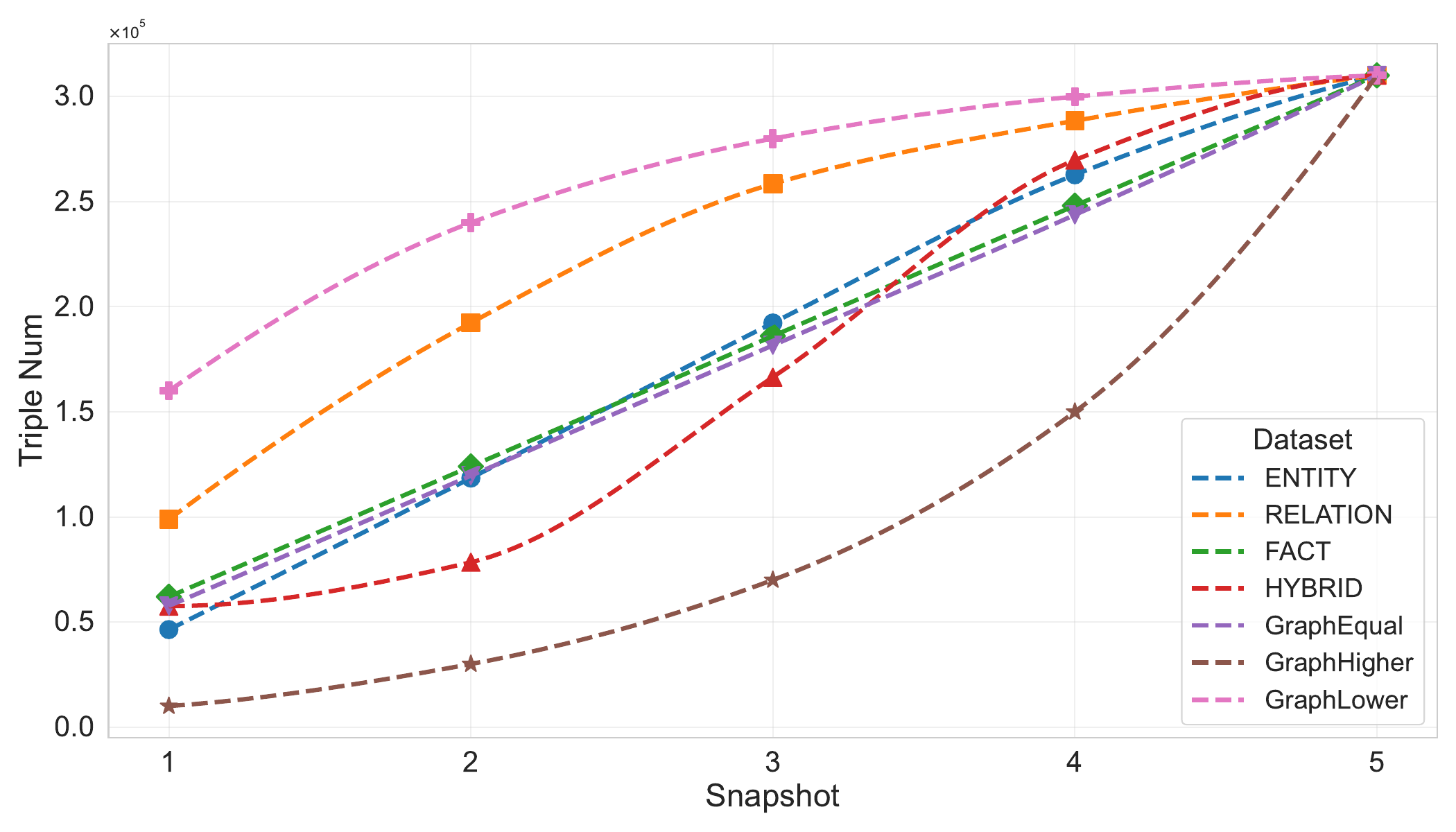}
  \vspace{-5mm} % 调整图像与标题之间的间距（负值减少间距）
  \caption{Number of Triples Across Snapshots for Dataset with Different Growth Strategies.}
  \label{fig5}
\end{figure}

\subsubsection{Datasets}
We have used seven publicly available datasets for our CKGE experiments, which include four datasets ENTITY, RELATION, FACT, and HYBRID~\cite{lkge}, and three additional datasets GraphEqual, GraphHigher, and GraphLower proposed by~\cite{incde}. ENTITY, RELATION, and FACT simulate the evolving scenarios where the number of entities, relations, and triples are uniformly increased, respectively. 
In contrast, HYBRID is designed to capture non-uniform growth patterns across entity, relation and triple dimensions, making it comparatively more challenging. 
% y et al. argue that the above four datasets are constructed in a way that has limitations, firstly that their construction method determines that all new ternary groups contain at least one existing entity, and secondly that the uniform growth method does not simulate the realistic scenarios well, so three new datasets are proposed, which are still disaggregated on DB15K-237, and 
To explore the impact varying scales of new triples on CKGE performances, GraphEqual is proposed to capture the uniform triples growth, while GraphHigher, and GraphLower simulate the ternary group growth rate increasing and decreasing, respectively. 
All above seven datasets all set the time step to 5, and the ratio of the number of training, validation and test sets is 3:1:1. Figure~\ref{fig5} shows the number of triples in different datasets during the five stages of continuous learning. More detailed statistics of these datasets can be found in Table \ref{t7}.
% As a method for scale adaptivity, we believe that the results on these datasets can well illustrate the superiority of our method.

\subsubsection{Compared Baselines}
We evaluate our approach against diverse baseline methods spanning fundamental approaches (snapshot-only training, fine-tuning with parameter inheritance), dynamic architecture-based methods (PNN, CWR)~\cite{pnn,lomonaco2017core50}, memory replay-based approaches (GEM, EMR, DiCGRL)~\cite{gem, wang2019sentence,kou2020disentangle}, regularization-based methods (SI, EWC)~\cite{zenke2017continual, kirkpatrick2017overcoming}, and recent CKGE-specific methods (LKGE, IncDE, FastKGE)\cite{lkge,incde,fastkge}.

\subsubsection{Metrics}

We evaluate the performance of our model on the link prediction task by replacing the head or tail entity of the test triples with all other entities, ranking the scores, and calculating metrics such as Mean Reciprocal Rank (MRR), Hits@1, and Hits@10, where higher values indicate better performance. For each snapshot $i$, we compute the average of these metrics over all test sets from snapshot $1$ to $i$, with the final results coming from the model of the last snapshot. 

To evaluate the model's ability to retain knowledge from earlier tasks, we propose a novel metric called Resistance to Forgetting (RtF). While the commonly used Backward Transfer (BWT)~\cite{gem} metric measures absolute performance degradation, we argue that relative performance changes provide a more objective assessment of forgetting. RtF is defined as:

\begin{equation}
RtF = \frac{1}{n-1} \sum_{i=1}^{n-1} \frac{h_{n,i}}{h_{i,i} + h_{n,i}}
\end{equation}

where \( n \) is the total number of snapshots, \( h_{i,i} \) is the MRR score on the test set of snapshot \( i \) immediately after training on snapshot \( i \), and \( h_{n,i} \) is the MRR score on the test set of snapshot \( i \) after training on all snapshots. The RtF metric evaluates the relative retention of previously learned knowledge by considering the ratio between final and peak performance. Higher RtF scores indicate better resistance to forgetting, with a value of \( 1 \) indicating perfect retention or positive transfer, and a value of \( 0 \) indicating complete forgetting. This formulation provides a more intuitive way to compare different models' capability in preserving historical knowledge.

\subsubsection{Implementation Details}
We conduct our experiments on NVIDIA L40 GPUs with the PyTorch framework. For fair comparison, all methods use TransE with the Adam optimizer. The training consists of 1 epoch for dimension expansion, using a replay sample size of $k=30$, and 200 epochs with early stopping for general training. The learning rate is 0.0001 with a margin loss of 8.0. The balance factor $\alpha$ is set to 0.01.
To ensure the generalizability of our method, we adopt a unified configuration across all datasets: the scaling factor $r=1.25$ and the dimensional increment step $step=10$ are fixed throughout. This consistency in hyperparameters avoids dataset-specific tuning and demonstrates the robustness of our approach under varying scales and graph structures.
All experimental results are averaged over 5 runs to ensure fairness. The results of the comparative methods are obtained from the corresponding original publications~\cite{lkge,incde,fastkge}.

\begin{table*}[htb!]
\centering
\setlength{\tabcolsep}{1mm}
\setlength{\abovecaptionskip}{10pt}
\caption{Main experimental results on ENTITY, RELATION, FACT, HYBRID, and GraphEqual datasets. Bold scores represent the best results, while underlined scores indicate the second-best performance across all methods.}
\resizebox{\textwidth}{!}{%
% \small
% \Large
\begin{tabular}{l ccc ccc ccc ccc ccc}
\hline
\multirow{3}{*}{Method} & \multicolumn{3}{c}{ENTITY} & \multicolumn{3}{c}{RELATION} & \multicolumn{3}{c}{FACT} & \multicolumn{3}{c}{HYBRID} & \multicolumn{3}{c}{GraphEqual} \\
 & MRR & H@1 & H@10 & MRR & H@1 & H@10 & MRR & H@1 & H@10 & MRR & H@1 & H@10 & MRR & H@1 & H@10 \\
\hline
Fine-tune & 0.165 & 0.085 & 0.321  & 0.093 & 0.039 & 0.195 & 0.172 & 0.090 & 0.339 & 0.135 & 0.069 & 0.262 & 0.183 & 0.096 & 0.358 \\
\hline
PNN & 0.229 & 0.130 & 0.425 & 0.167 & 0.096 & 0.305 & 0.157 & 0.084 & 0.290 & 0.185 & 0.101 & 0.349 & 0.212 & 0.118 & 0.405 \\
CWR & 0.088 & 0.028 & 0.202 & 0.021 & 0.010 & 0.043 & 0.083 & 0.030 & 0.192 & 0.037 & 0.015 & 0.077 & 0.122 & 0.041 & 0.277 \\
\hline
GEM & 0.165 & 0.085 & 0.321 & 0.093 & 0.040 & 0.196 & 0.175 & 0.092 & 0.345 & 0.136 & 0.070 & 0.263 & 0.189 & 0.099 & 0.372  \\
EMR & 0.171 & 0.090 & 0.330 & 0.111 & 0.052 & 0.225 & 0.171 & 0.090 & 0.337 & 0.141 & 0.073 & 0.267 & 0.185 & 0.099 & 0.359 \\
DiCGRL & 0.107 & 0.057 & 0.211 & 0.133 & 0.079 & 0.241 & 0.162 & 0.084 & 0.320 & 0.149 & 0.083 & 0.277 & 0.104 & 0.040 & 0.226 \\
\hline
SI & 0.154 & 0.072 & 0.311 & 0.113 & 0.055 & 0.224 & 0.172 & 0.088 & 0.343 & 0.111 & 0.049 & 0.229 & 0.179 & 0.092 & 0.353 \\
EWC & 0.229 & 0.130 & 0.423 & 0.165 & 0.093 & 0.306 & 0.201 & 0.113 & 0.382 & 0.186 & 0.102 & 0.350 & 0.207 & 0.113 & 0.400 \\
\hline
LKGE & 0.234 & 0.136 & 0.425 & 0.192 & 0.106 & 0.366 & 0.210 & 0.122 & 0.387 & 0.207 & 0.121 & 0.379 & 0.214 & 0.118 & 0.407 \\
FastKGE & 0.239 & 0.146 & 0.427 & 0.185 & 0.107 & 0.359 & 0.203 & 0.117 & 0.384 & 0.211 & 0.128 & 0.382 & - & - & - \\
IncDE & \underline{0.253} & \underline{0.151} & \underline{0.448} & \underline{0.199} & \underline{0.111} & \underline{0.370} & \underline{0.216} & \underline{0.128} & \underline{0.391} & \underline{0.224} & \underline{0.131} & \textbf{0.401} & \underline{0.234} & \underline{0.134} & \underline{0.432} \\
\hline
\textbf{SAGE} & \textbf{0.280} & \textbf{0.176} & \textbf{0.477} & \textbf{0.217} & \textbf{0.122} & \textbf{0.397} & \textbf{0.222} & \textbf{0.134} & \textbf{0.394} & \textbf{0.224} & \textbf{0.131} & \underline{0.400} & \textbf{0.247} & \textbf{0.147} & \textbf{0.446} \\
\hline
\end{tabular}}
\label{t1}
\end{table*}

\begin{table}[htb!]
\centering
\setlength{\tabcolsep}{1mm}
\setlength{\abovecaptionskip}{10pt} 
\caption{Main experimental results on GraphHigher and GraphLower.}
\resizebox{\columnwidth}{!}{%
% \small
% \Large
\begin{tabular}{l ccc ccc}
\hline
\multirow{3}{*}{Method} & \multicolumn{3}{c}{GraphHigher} & \multicolumn{3}{c}{GraphLower} \\
 & MRR & H@1 & H@10 & MRR & H@1 & H@10 \\
\hline
Fine-tune & 0.198 & 0.108 & 0.375 & 0.185 & 0.098 & 0.363 \\
\hline
PNN & 0.186 & 0.097 & 0.364 & 0.213 & 0.119 & 0.407 \\
CWR & 0.189 & 0.096 & 0.374 & 0.032 & 0.005 & 0.080 \\
\hline
GEM & 0.197 & 0.109 & 0.372 & 0.170 & 0.084 & 0.346 \\
EMR & 0.202 & 0.113 & 0.379 & 0.188 & 0.101 & 0.362 \\
DiCGRL & 0.116 & 0.041 & 0.242 & 0.102 & 0.039 & 0.222 \\
\hline
SI & 0.190 & 0.099 & 0.371 & 0.186 & 0.099 & 0.366 \\
EWC & 0.198 & 0.106 & 0.385 & 0.210 & 0.116 & 0.405 \\
\hline
LKGE & 0.207 & 0.120 & 0.382 & 0.210 & 0.116 & 0.403 \\
IncDE & \underline{0.227} & \underline{0.132} & \underline{0.412} & \underline{0.228} & \underline{0.129} & \underline{0.426} \\
\hline
\textbf{SAGE} & \textbf{0.239} & \textbf{0.143} & \textbf{0.426} & \textbf{0.237} & \textbf{0.138} & \textbf{0.432} \\
\hline
\end{tabular}}
\vspace{-10pt} % 手动减少间距
\label{t2}
\end{table}

\begin{table}[htb!]
\centering
\setlength{\tabcolsep}{1mm}
\setlength{\abovecaptionskip}{10pt} 
\caption{Ablation study results on GraphHigher and GraphLower. "w/o" indicates the removal of specific components: Scale Estimate (SE), Difficulty-based Sampling (DS), Lightweight Expansion (LE), and Dynamic Integration (DI).}
\resizebox{\columnwidth}{!}{%
% \small
% \Large
\begin{tabular}{l ccc ccc}
\hline
\multirow{3}{*}{Method} & \multicolumn{3}{c}{GraphHigher} & \multicolumn{3}{c}{GraphLower} \\
 & MRR & H@1 & H@10 & MRR & H@1 & H@10 \\
\hline
SAGE & \textbf{0.239} & \textbf{0.143} & \textbf{0.426} & \textbf{0.237} & \textbf{0.138} & \textbf{0.432} \\
w/o SE & 0.237 & 0.143 & 0.420 & 0.218 & 0.126 & 0.404 \\
w/o DS & 0.238 & 0.144 & 0.426 & 0.230 & 0.127 & 0.417 \\
w/o LE & 0.236 & 0.142 & 0.423 & 0.199 & 0.111 & 0.371 \\
w/o DI & 0.237 & 0.143 & 0.424 & 0.205 & 0.114 & 0.383 \\
\hline
\end{tabular}}
\vspace{-10pt} % 手动减少间距
\label{t3}
\end{table}

\subsection{Main Results}
Table~\ref{t1} and ~\ref{t2} present the performance of SAGE and other baseline methods on continual knowledge graph embedding tasks across different benchmark datasets.

The experimental results demonstrate significant improvements across all evaluation metrics. Compared to conventional methods (Fine-tune, PNN, CWR, GEM, EMR, DiCGRL, SI, EWC), SAGE improves MRR by 4.2\%-20.5\%, H@1 by 2.9\%-14.8\%, and H@10 by 4.6\%-35.4\%. The substantial performance gaps across all metrics indicate that these conventional methods have not effectively addressed the core challenges in continuous knowledge graph embedding tasks, particularly in balancing the trade-off between maintaining historical knowledge and adapting to new information.

When compared to contemporary state-of-the-art methods(LKGE, IncDE, FastKGE), SAGE achieves either the best or second-best performance across most datasets. Averaging over seven datasets, SAGE shows improvements of 1.38\%, 1.25\%, and 1.6\% in MRR, H@1, and H@10 respectively, demonstrating its consistent advantage.

Finally, we analyze the performance of our method across different knowledge graph evolution patterns. In conventional expansion patterns (Table~\ref{t1}), SAGE demonstrates particularly strong performance in entity-centric and relation-centric evolution, with notable improvements of 2.7\% and 1.8\% in MRR respectively,validating that our adaptive dimension expansion mechanism effectively balances lightweight expansion with model adaptation.
For fact-based evolution, SAGE achieves the best performance, although the lead is relatively marginal, suggesting that there is still room for further optimization.
In the more complex HYBRID scenario that combines multiple evolution patterns, SAGE matches or slightly outperforms state-of-the-art methods. 
More significantly, when tested on more challenging growth patterns (Table~\ref{t2}, GraphHigher and GraphLower), SAGE exhibits even more pronounced advantages, indicating its robust capability in handling complex evolution scenarios with varying growth rates.
 These results validate the effectiveness of SAGE for continuous knowledge graph embedding.

\subsection{Ablation Study}
To systematically evaluate the contribution of each component in our framework, we conduct ablation studies on the GraphHigher and GraphLower datasets, which exhibit particularly challenging evolution patterns with substantial changes in graph scale. Results on the other five datasets are included in Table~\ref{t6}.

\paragraph{Ablation Settings.}
We define the following variants to isolate the impact of each design choice:

\begin{itemize}
    \item \textbf{w/o SE (Scale Estimation)}: The model starts with a fixed dimension of 200 and increases it by a constant step size of 10 for each new snapshot, independent of specific graph characteristics. This removes the scale-aware guidance mechanism.
    \item \textbf{w/o DS (Difficulty-based Sampling)}: The replay buffer is filled via random sampling of past triplets, rather than focusing on more informative or harder examples.
    \item \textbf{w/o LE (Lightweight Expansion)}: The model adopts an Incremental Only strategy by directly training parameters of newly added dimensions using replay samples, without leveraging old representations through transfer or mapping.
    \item \textbf{w/o DI (Dynamic Integration)}: During training on new triples, the model does not apply footprint-based distillation weights (computed from frequency and representation quality) to modulate learning of old entities and relations. Instead, all embeddings are updated uniformly, which can disrupt existing knowledge and lead to catastrophic forgetting.
\end{itemize}

\paragraph{Ablation Results.}
As shown in Table~\ref{t3}, the LE and DI components contribute most significantly to performance gains. On GraphLower, removing LE or DI causes MRR drops of 0.038 and 0.032 respectively, underscoring the importance of preserving and reusing prior knowledge during embedding expansion. The SE mechanism yields moderate but consistent improvements, especially in conservative expansion settings such as GraphLower (-0.019 MRR). DS also offers stable gains by focusing replay on informative samples, enhancing convergence and robustness.

\begin{figure}[!t]
  \centering
  \includegraphics[width=\linewidth]{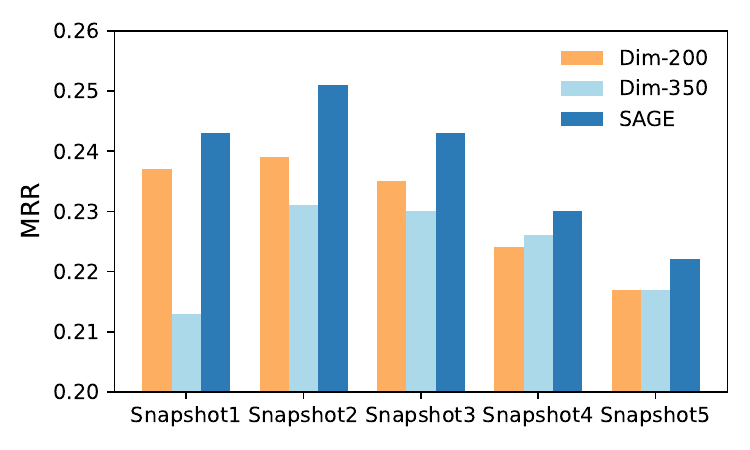}
  \vspace{-5mm} % 调整图像与标题之间的间距（负值减少间距）
  \caption{Performance comparison of SAGE with different dimension settings over snapshots on FACT.}
  \label{fig6}
\end{figure}

\begin{figure}[!t]
  \centering
  \includegraphics[width=\linewidth]{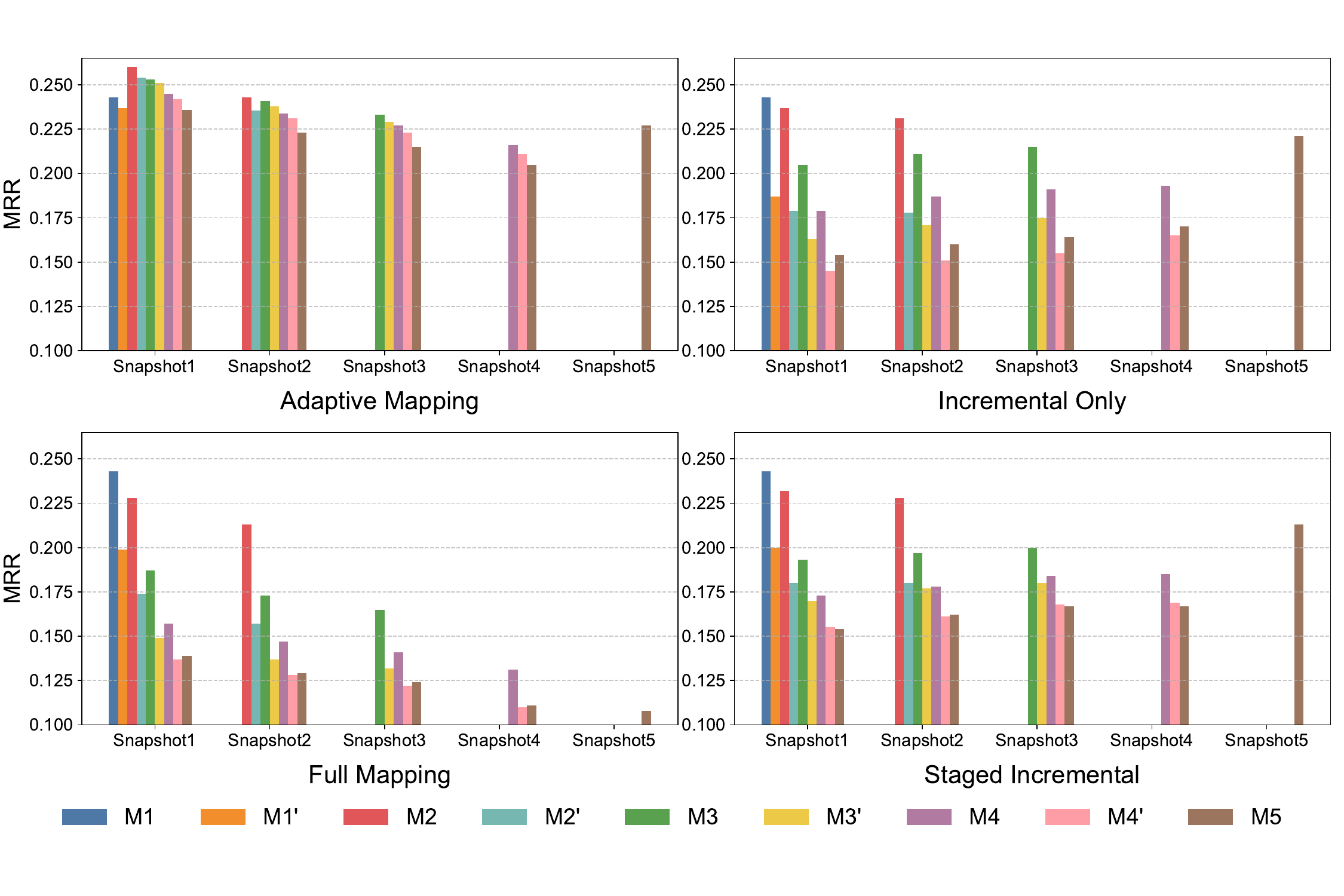}
  \vspace{-5mm} % 调整图像与标题之间的间距（负值减少间距）
  \caption{Performance comparison of different dimension expansion methods across snapshots. Mi represents the model's performance on queries from the i-th stage using original dimensions, while Mi' shows the performance using expanded dimensions learned during stage i+1.}
  \label{fig7}
\end{figure}
\subsection{Adaptive versus Static Dimensions}

To further validate the effectiveness of our adaptive expansion strategy, we compare SAGE with two fixed-dimension configurations: \textbf{Dim-200}, a standard setting used in prior works, and \textbf{Dim-350}, which approximates the upper bound of our expanded dimensions across all snapshots. As shown in Figure~\ref{fig6}, these static configurations exhibit fluctuating performance across different time steps. Dim-200 often lacks sufficient capacity to represent increasingly complex knowledge graphs in later stages, while Dim-350 fails to consistently outperform it despite offering higher dimensional expressiveness.

This phenomenon highlights a critical observation: \emph{higher embedding dimensions do not necessarily lead to better performance}, particularly in the early stages of KG evolution when data volume is limited. In such scenarios, excessive dimensionality can lead to underfitting, as the model is unable to effectively utilize the surplus capacity due to insufficient training signal. This mismatch between model size and data complexity not only results in wasted parameters but can also destabilize the training process.

By contrast, SAGE dynamically adjusts the embedding size in accordance with the observed growth of the knowledge graph, allocating dimensions gradually as more data becomes available. This adaptive mechanism ensures a better balance between representation power and data availability, enabling efficient learning across all stages. As demonstrated in Figure~\ref{fig6}, SAGE consistently achieves the best performance throughout the continual learning process, confirming its robustness in maintaining alignment between model capacity and task complexity.

% To validate the superiority of SAGE over fixed dimension approaches, we compare it with two settings:  Dim-200 (standard dimension used in previous work) and Dim-350 (upper bound of expanded dimensions). As shown in Figure~\ref{fig6} , while Dim-350 and Dim-200 show varying performance across different snapshots with inconsistent advantages over each other, SAGE consistently achieves the best results in all five stages. This demonstrates that our adaptive dimension strategy is more effective than static dimension settings, regardless of whether they are set high or low. The consistent optimal performance across all snapshots further validates SAGE's robust adaptation capability throughout the continual learning process.

\subsection{Analysis of Expansion Strategies}

To validate the effectiveness of our embedding expansion strategy, we compare our approach against several representative alternatives that reflect different principles for handling newly introduced dimensions. Specifically, we evaluate four strategies:

\begin{itemize}
    \item \textbf{Incremental Only:} The model trains only the newly added dimensions, while keeping the original embeddings fixed. This represents a naive strategy without any integration of prior knowledge.
    
    \item \textbf{Staged Incremental:} The new dimensions are added in multiple rounds following the predefined step size. Parameters are optimized layer-by-layer to gradually increase capacity, aiming to reduce sudden representation shifts.
    
    \item \textbf{Full Mapping:} A single transformation network maps the entire original representation to the expanded dimensional space. Both the original and new dimensions are updated jointly during training, which may introduce interference with prior knowledge.
    
    \item \textbf{Adaptive Mapping (ours):} A lightweight neural network is used to map the original embedding to the additional dimensions, which are then concatenated to the original representation. Only the newly added dimensions are updated, preserving prior knowledge while enabling flexible capacity expansion.
\end{itemize}

This setup allows us to systematically compare whether incorporating original representations in the training process improves the stability and effectiveness of dimension expansion, especially in continual knowledge graph settings.

The experimental results demonstrate the advantages of our Adaptive Mapping approach. As shown in Figure~\ref{fig7}, Adaptive Mapping maintains consistently higher MRR scores across different snapshots compared to other methods. While Full Mapping shows comparable initial performance, its performance degrades more severely in later snapshots, suggesting potential interference with previously learned representations. Both Incremental Only and Staged Incremental methods exhibit lower performance, indicating the importance of learning the relationships between original and expanded dimensions rather than simply optimizing new parameters. Moreover, our method shows better stability in maintaining performance on historical queries (e.g., M1 and M2) while accommodating new data.

\subsection{Analysis of Replay Sample Size}

\begin{table}[t]
\centering
\setlength{\tabcolsep}{1.5mm}
\setlength{\abovecaptionskip}{10pt}
\caption{Performance comparison with different replay sample sizes ($k$).}
\resizebox{\columnwidth}{!}{%
\begin{tabular}{ccccccc}
\hline
 & \multicolumn{3}{c}{GraphHigher} & \multicolumn{3}{c}{GraphLower} \\
k       & MRR & H@1 & H@10 & MRR & H@1 & H@10 \\
\hline
10      & 0.237 & 0.142 & 0.422 & 0.236 & 0.137 & 0.430 \\
30      & 0.239 & 0.143 & 0.426 & 0.237 & 0.138 & 0.432 \\
60      & 0.239 & 0.143 & 0.425 & 0.237 & 0.138 & 0.432 \\
100     & 0.238 & 0.143 & 0.424 & 0.237 & 0.139 & 0.433 \\
1000    & 0.235 & 0.142 & 0.414 & 0.239 & 0.141 & 0.435 \\
5000    & 0.232 & 0.141 & 0.411 & 0.233 & 0.136 & 0.424 \\
\hline
\end{tabular}}
\label{t4}
\end{table}

We analyze the impact of replay sample size ($k$) on model performance. As shown in Table~\ref{t4}, the performance remains generally stable across different $k$ values (from 10 to 5000), with optimal results achieved using relatively small $k$ values (30-100). Interestingly, GraphLower shows a slight tendency to benefit from larger $k$ values (as seen in the marginal improvements with $k$=1000), which can be attributed to its slower growth rate leading to more triplets in early stages. In contrast, GraphHigher maintains more consistent performance across different $k$ values, suggesting it can effectively preserve knowledge with minimal replay samples. This robustness to replay sample size stems from the model’s architectural design: instead of retraining the full embedding space, which would involve a large number of parameters, we employ a lightweight transformation module to project old representations into the expanded space. This parameter-efficient transfer mechanism mitigates the dependence on large replay memories by enabling effective knowledge retention with limited historical data. These results demonstrate that our dimension expansion strategy can maintain effectiveness with a small replay memory.

\subsection{Analysis of Computational Efficiency}

\begin{table}[t]
\centering
\setlength{\tabcolsep}{1.5mm}
\setlength{\abovecaptionskip}{10pt}
\renewcommand{\arraystretch}{1.15}
\caption{Estimated training cost (FLOPs) of different modules.}
\label{tab:flops}
\resizebox{\columnwidth}{!}{%
\begin{tabular}{llcc}
\hline
Method & Module & FACT & RELATION \\
\hline
Fixed-Dim & LKGE / IncDE & $1.17 \times 10^{14}$ & $1.20 \times 10^{14}$ \\
\hline
\multirow{2}{*}{SAGE} & Dim-expand & $1.50 \times 10^{9}$ & $5.63 \times 10^{8}$ \\
                      & Integration & $1.35 \times 10^{14}$ & $1.80 \times 10^{14}$ \\
\hline
\end{tabular}}
\end{table}

We analyze the training cost of SAGE using a FLOPs-based estimation~\cite{hoffmann2022training}. As shown in Table~\ref{tab:flops}, the computational overhead of the \textit{Dim-expand} module is nearly negligible, with FLOPs on the order of $10^9$, which is minimal compared to other components. In contrast, the \textit{Integration} module introduces higher overhead due to additional trainable parameters, especially under high-dimensional settings. Nevertheless, the total cost remains controllable and comparable to fixed-dimension baselines, demonstrating that our dynamic design achieves a good balance between scalability and efficiency.

\subsection{Analysis of Forgetting Resistance}

We evaluate our method on four datasets: ENTITY, RELATION, FACT and HYBRID. The baseline methods include traditional continual learning approaches (GEM based on replay, CWR based on dynamic architecture, and EWC based on regularization), as well as state-of-the-art methods specifically designed for evolving knowledge graphs (LKGE and IncDE).
We examine their performance with Resistance to Forgetting (RtF) metric, where RtF=0.5 represents the theoretical upper bound indicating complete resistance to catastrophic forgetting. As shown in Table~\ref{t5}, several methods including SAGE, EWC, and LKGE achieve near-optimal RtF scores approaching the 0.5 threshold, demonstrating strong capabilities in preserving previously learned knowledge.
While methods like EWC and LKGE show slightly better scores in some cases, SAGE achieves these high levels of forgetting resistance while simultaneously maintaining strong performance in knowledge acquisition, as demonstrated in previous experiments.

\begin{table}[t]
\centering
\setlength{\tabcolsep}{1.5mm}
\setlength{\abovecaptionskip}{10pt}
\caption{Comparison of Resistance to Forgetting (RtF) across different methods and datasets. Pink highlights indicate better performance in the respective categories.}
\resizebox{\columnwidth}{!}{%
\begin{tabular}{lcccccc}
\hline
RtF      & CWR          & GEM          & EWC          & IncDE        & LKGE         & SAGE        \\
\hline
ENTITY   & 0.395        & 0.448        & \cellcolor{pink!25}0.499 & 0.404        & \cellcolor{pink!25}0.497 & \cellcolor{pink!25}0.499 \\
RELATION & 0.400        & 0.449        & \cellcolor{pink!25}0.499 & 0.408        & \cellcolor{pink!25}0.499 & \cellcolor{pink!25}0.485        \\
FACT     & 0.435        & 0.475        & \cellcolor{pink!25}0.499 & 0.413        & \cellcolor{pink!25}0.492        & \cellcolor{pink!25}0.494        \\
HYBRID   & 0.398        & \cellcolor{pink!25}0.496 & \cellcolor{pink!25}0.499 & 0.429        & \cellcolor{pink!25}0.496 & \cellcolor{pink!25}0.489        \\
\hline
\end{tabular}}
\label{t5}
\end{table}

\section{Conclusion} 
In this paper, we proposed SAGE (Scale-Aware Gradual Evolution), a framework for continual knowledge graph embedding (CKGE) that dynamically adapts to evolving knowledge graphs. SAGE introduces Scale Estimation, Lightweight Embedding Expansion, and Dynamic Distillation to balance knowledge retention and the integration of new information.
Extensive experiments on seven benchmarks demonstrate that SAGE consistently outperforms state-of-the-art methods, achieving optimal performance at every stage of knowledge graph evolution. Its adaptive embedding strategy proves superior to static approaches, while its strong Resistance to Forgetting (RtF) highlights its robustness in preserving historical knowledge.
SAGE provides an effective and scalable solution for CKGE, offering insights for adaptive learning in dynamic settings and opening opportunities for broader applications in evolving data environments.

\begin{acks}
This work was supported by the National Key Research and Development Program of China (2022YFC3303600), the National Natural Science Foundation of China (No. 62137002, 62293550, 62293553, 62477037, 62176207, 62192781), “LENOVO-XJTU” Intelligent Industry Joint Laboratory Project, the Shaanxi Undergraduate and Higher Education Teaching Reform Research Program (No. 23BY195), the Xi’an Jiaotong University City College Research Project (No. 2024Y01), and the Zhongguancun Academy Project (No. 20240103).
\end{acks}

\bibliographystyle{ACM-Reference-Format}
\balance
\bibliography{ref}

\appendix

\section{Related Work}

Continual Knowledge Graph Embedding (CKGE) addresses the challenge of enabling models to incorporate new knowledge while retaining previously learned information, which traditional knowledge graph embedding (KGE) methods such as TransE~\cite{bordes2013translating}, DistMult~\cite{yang2014embedding}, and ComplEx~\cite{trouillon2016complex} fail to handle efficiently. Unlike static KGE approaches, CKGE methods aim to incrementally learn from evolving knowledge graphs, avoiding retraining from scratch and mitigating catastrophic forgetting. Existing CKGE methods can be categorized into three main groups.

The first category includes dynamic architecture-based methods~\cite{rusu2016progressive,lomonaco2017core50}, which dynamically adjust neural architectures to accommodate new knowledge while preserving old parameters. These methods increase model capacity as new data arrives but face challenges in maintaining efficiency. The second category consists of memory replay-based methods~\cite{gem, kou2020disentangle}, which retain old knowledge by replaying stored samples of past data, ensuring that the model does not forget previously learned embeddings. Lastly, regularization-based methods~\cite{zenke2017continual, kirkpatrick2017overcoming} impose constraints on weight updates to preserve critical information from earlier tasks, effectively alleviating catastrophic forgetting. While these approaches have shown promise, they often overlook the importance of learning new knowledge in an appropriate order for graph data and struggle to balance the integration of new and old knowledge effectively. Additionally, existing CKGE datasets~\cite{hamaguchi2017knowledge, kou2020disentangle, daruna2021continual} often assume that new triples contain at least one old entity, which is unrealistic for real-world evolving knowledge graphs like DBpedia~\cite{lehmann2015dbpedia}.

Several baselines have been proposed to address the unique challenges of CKGE. LKGE\cite{lkge} incorporates temporal information to capture evolving knowledge graph dynamics. IncDE\cite{incde} uses incremental decomposition to efficiently update embeddings while preserving previously learned knowledge. FastKGE~\cite{fastkge} focuses on scalable updates, enabling faster adaptation to large-scale graphs. Recent work has revisited the evaluation protocols and benchmarks in CKGE~\cite{zhao2025rethinking} identify limitations in existing datasets and propose more realistic benchmarks, providing a comprehensive analysis of current CKGE methods under continual learning scenarios.
These baselines demonstrate the potential of CKGE methods to handle real-world, dynamically evolving knowledge graphs.

Beyond these baselines, recent research has explored related directions to enhance CKGE. For instance, memory-efficient techniques like low-rank adapters (LoRAs)\cite{hu2022lora, zhang2023llama} have been applied in large language models (LLMs) to fine-tune them efficiently, and similar ideas are being adapted to CKGE\cite{wang2023lora} to alleviate catastrophic forgetting. Parameter-efficient approaches, such as learning small sets of reserved entities to represent all entities~\cite{ding2023parameter}, have also been proposed to improve scalability. Furthermore, temporal KGE models~\cite{kazemi2018simple, pan2021hyperbolic, shang2023askrl} explicitly capture time-sensitive patterns, complementing CKGE for evolving knowledge graphs. These advancements highlight the growing interest in designing efficient, scalable, and robust CKGE models for real-world applications.

Beyond the direct scope of CKGE, a broader range of research explores theoretical and practical aspects of model scalability, parameter growth, and evolving systems, which are highly relevant to the continual learning and embedding space. For instance, studies on scaling laws\cite{kaplan2020scaling, henighan2020scaling} reveal that performance improvements often follow predictable patterns as model size, dataset size, or computational resources increase. These findings provide insights into designing scalable embeddings for knowledge graphs. Moreover, the sparsity hypothesis\cite{luo2024sparsing, evci2020rigging} shows that neural networks often operate effectively in sparse parameter spaces, which is particularly aligned with knowledge graphs' inherent sparsity. Leveraging such sparsity can improve computational efficiency and allow for scalable CKGE models. Additionally, research on evolving neural systems\cite{xu2024interactive, huang2024evochart, xu2025phi} and dynamic graph representation learning\cite{kazemi2020representation, pareja2020evolvegcn} explores how models can grow and adapt in response to changing data, offering methodologies for long-term learning in dynamic environments. These works, though not directly tied to CKGE, provide theoretical and practical foundations for addressing challenges in scalability and adaptability, enabling CKGE research to progress toward more efficient and robust solutions for evolving knowledge graphs.

\section{Supplementary Tables}

\begin{table*}[h!]
\centering
\setlength{\tabcolsep}{1mm}
\setlength{\abovecaptionskip}{10pt} % 表格与注释之间的距离
\resizebox{\textwidth}{!}{%
% \small
% \Large
\begin{tabular}{l ccc ccc ccc ccc ccc}
\hline
\multirow{3}{*}{Method} & \multicolumn{3}{c}{ENTITY} & \multicolumn{3}{c}{RELATION} & \multicolumn{3}{c}{FACT} & \multicolumn{3}{c}{HYBRID} & \multicolumn{3}{c}{GraphEqual} \\
 & MRR & H@1 & H@10 & MRR & H@1 & H@10 & MRR & H@1 & H@10 & MRR & H@1 & H@10 & MRR & H@1 & H@10 \\
\hline
SAGE & \textbf{0.28} & \textbf{0.176} & \textbf{0.477} & \textbf{0.217} & \textbf{0.122} & \textbf{0.399} & \textbf{0.222} & \textbf{0.135} & \textbf{0.396} & \textbf{0.224} & \textbf{0.131} & \textbf{0.4} & \textbf{0.247} & \textbf{0.147} & \textbf{0.446} \\
w/o SE & 0.268 & 0.158 & 0.458 & 0.201 & 0.113 & 0.375 & 0.212 & 0.127 & 0.381 & 0.214 & 0.122 & 0.39 & 0.232 & 0.135 & 0.423 \\
w/o DS & 0.277 & 0.143 & 0.474 & 0.199 & 0.111 & 0.37 & 0.22 & 0.134 & 0.393 & 0.222 & 0.13 & 0.397 & 0.244 & 0.145 & 0.44 \\
w/o LE & 0.199 & 0.12 & 0.346 & 0.034 & 0.012 & 0.065 & 0.172 & 0.104 & 0.305 & 0.161 & 0.081 & 0.305 & 0.206 & 0.115 & 0.388 \\
w/o DI & 0.212 & 0.129 & 0.368 & 0.065 & 0.03 & 0.122 & 0.187 & 0.108 & 0.347 & 0.171 & 0.089 & 0.321 & 0.212 & 0.12 & 0.397 \\
\hline
\end{tabular}}
\caption{Ablation study results on ENTITY, RELATION, FACT, HYBRID, and GraphEqual. "w/o" indicates the removal of specific components: Scale Estimate (SE), Difficulty-based Sampling (DS), Lightweight Expansion (LE), and Dynamic Integration (DI).}
\label{t6}
\end{table*}
\begin{table*}[h!]
\centering
\setlength{\tabcolsep}{1mm}
\setlength{\abovecaptionskip}{10pt}
\resizebox{\textwidth}{!}{%
\begin{tabular}{l ccc ccc ccc ccc ccc}
\hline
\multirow{2}{*}{Dataset} & \multicolumn{3}{c}{Snapshot 1} & \multicolumn{3}{c}{Snapshot 2} & \multicolumn{3}{c}{Snapshot 3} & \multicolumn{3}{c}{Snapshot 4} & \multicolumn{3}{c}{Snapshot 5} \\
 & $N_E$ & $N_R$ & $N_T$ & $N_E$ & $N_R$ & $N_T$ & $N_E$ & $N_R$ & $N_T$ & $N_E$ & $N_R$ & $N_T$ & $N_E$ & $N_R$ & $N_T$ \\
\hline
ENTITY & 2,909 & 233 & 46,388 & 5,817 & 236 & 72,111 & 8,275 & 236 & 73,785 & 11,633 & 237 & 70,506 & 14,541 & 237 & 47,326 \\
RELATION & 11,560 & 48 & 98,819 & 13,343 & 96 & 93,535 & 13,754 & 143 & 66,136 & 14,387 & 190 & 30,032 & 14,541 & 237 & 21,594 \\
FACT & 10,513 & 237 & 62,024 & 12,779 & 237 & 62,023 & 13,586 & 237 & 62,023 & 13,894 & 237 & 62,023 & 14,541 & 237 & 62,023 \\
HYBRID & 8,628 & 86 & 57,561 & 10,040 & 102 & 20,873 & 12,779 & 151 & 88,017 & 14,393 & 209 & 103,339 & 14,541 & 237 & 40,326 \\
GraphEqual & 2,908 & 226 & 57,636 & 5,816 & 235 & 62,023 & 8,724 & 237 & 62,023 & 11,632 & 237 & 62,023 & 14,541 & 237 & 66,411 \\
GraphHigher & 900 & 197 & 10,000 & 1,838 & 221 & 20,000 & 3,714 & 234 & 40,000 & 7,467 & 237 & 80,000 & 14,541 & 237 & 160,116 \\
GraphLower & 7,505 & 237 & 160,000 & 11,258 & 237 & 80,000 & 13,134 & 237 & 40,000 & 14,072 & 237 & 20,000 & 14,541 & 237 & 10,116 \\
\hline
\end{tabular}}
\caption{The statistics of datasets. $N_E$, $N_R$ and $N_T$ denote the number of cumulative entities, cumulative relations and current triples at each time $i$.}
\label{t7}
\end{table*}

The supplementary experiments evaluate the contribution of key components in our method, including Scale Estimate (SE), Difficulty-based Sampling (DS), Lightweight Expansion (LE), and Dynamic Integration (DI), by removing each component individually. The results demonstrate that the complete SAGE model consistently achieves the best performance across all datasets (ENTITY, RELATION, FACT, HYBRID, and GraphEqual) on all metrics (MRR, H@1, H@10). In contrast, removing any single component results in a noticeable drop in performance, underscoring the critical role of each module in the overall framework. Furthermore, the complete model exhibits significant advantages in more complex scenarios, such as HYBRID and GraphEqual, highlighting its robustness and effectiveness in handling diverse tasks. These findings confirm the superiority and soundness of our method's design.

\end{document}